\documentclass[english,conference]{IEEEtran}
\usepackage[latin9]{inputenc}
\usepackage{color}
\usepackage{babel}
\usepackage{array}
\usepackage{float}
\usepackage{units}
\usepackage{multirow}
\usepackage{algorithm2e}
\usepackage{amstext}
\usepackage{amssymb}
\usepackage{graphicx}
\PassOptionsToPackage{normalem}{ulem}
\usepackage{ulem}
\usepackage[unicode=true,pdfusetitle,
 bookmarks=false,
 breaklinks=false,pdfborder={0 0 1},backref=false,colorlinks=true]
 {hyperref}

\makeatletter

\providecommand{\tabularnewline}{\\}

\usepackage{cite}
\usepackage{amsmath,amssymb,amsfonts}
\usepackage{algorithmic}
\usepackage{graphicx}
\usepackage{textcomp}
\usepackage{xcolor}
\usepackage{balance}

\usepackage{colortbl} 
\definecolor{header_color}{rgb}{0.74,0.88,0.91}
\definecolor{even_color}{rgb}{0.9,0.9,0.9}
\definecolor{subheader_color}{rgb}{0.85,0.93,0.95}
\definecolor{childheader_color}{rgb}{1.0,0.93,0.87}

\@ifundefined{showcaptionsetup}{}{%
 \PassOptionsToPackage{caption=false}{subfig}}
\usepackage{subfig}
\makeatother

\begin{document}
\IEEEoverridecommandlockouts 
\def\BibTeX{{\rm B\kern-.05em{\sc i\kern-.025em b}\kern-.08em     
T\kern-.1667em\lower.7ex\hbox{E}\kern-.125emX}} 

\title{Generating Realistic Tabular Data with Large Language Models}

\author{Dang Nguyen, Sunil Gupta, Kien Do, Thin Nguyen, Svetha Venkatesh\\
Applied Artificial Intelligence Institute (A\textsuperscript{2}I\textsuperscript{2}),
Deakin University, Geelong, Australia\\
\textit{\{d.nguyen, sunil.gupta, k.do, thin.nguyen, svetha.venkatesh\}@deakin.edu.au}}
\maketitle
\begin{abstract}
While most generative models show achievements in image data generation,
few are developed for tabular data generation. Recently, due to success
of large language models (LLM) in diverse tasks, they have also been
used for tabular data generation. However, these methods do not capture
the correct correlation between the features and the target variable,
hindering their applications in downstream predictive tasks. To address
this problem, we propose a LLM-based method with three important improvements
to correctly capture the ground-truth feature-class correlation in
the real data. First, we propose a novel permutation strategy for
the input data in the fine-tuning phase. Second, we propose a feature-conditional
sampling approach to generate synthetic samples. Finally, we generate
the labels by constructing prompts based on the generated samples
to query our fine-tuned LLM. Our extensive experiments show that our
method significantly outperforms 10 SOTA baselines on 20 datasets
in downstream tasks. It also produces highly realistic synthetic samples
in terms of quality and diversity. More importantly, classifiers trained
with our synthetic data can even compete with classifiers trained
with the original data on half of the benchmark datasets, which is
a significant achievement in tabular data generation.
\end{abstract}

\section{Introduction\label{sec:Introduction}}

Recently, \textit{tabular data generation} has attracted a significant
attention from the research community as synthetic data can improve
different aspects of training data such as privacy \cite{Torfi2022},
quality \cite{Waheed2021}, fairness \cite{Rajabi2022}, and availability
\cite{Moon2020}. Particularly noteworthy is its capacity to overcome
usage restrictions while preserving privacy e.g.\textit{ synthetic
data replace real data in training machine learning (ML) models} \cite{Xu2019,Borisov2023}.

\begin{figure}[th]
\begin{centering}
\includegraphics[scale=0.45]{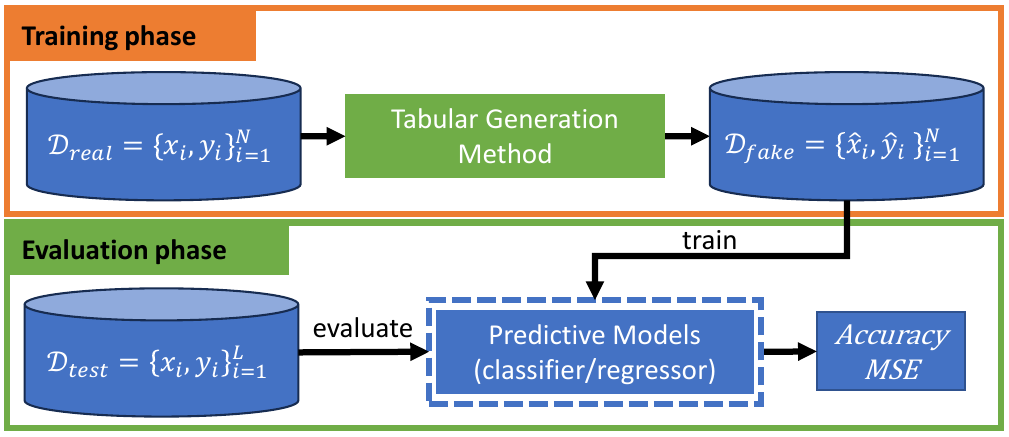}
\par\end{centering}
\caption{\label{fig:Training-and-evaluation}Training and evaluation of a tabular
generation method. \textbf{Training phase:} the tabular generation
method learns to generate a synthetic dataset ${\cal D}_{fake}$ to
approximate the real dataset ${\cal D}_{real}$. \textbf{Evaluation
phase:} ${\cal D}_{fake}$ is used to train a predictive model (e.g.
XGBoost). The predictive model is evaluated on a held-out real test
set ${\cal D}_{test}$ to compute accuracy or mean squared error (MSE).
\textit{A better score implies a better tabular generation method}.}
\end{figure}

To evaluate an image generation method, we often generate synthetic
images $\hat{x}$, and visualize $\hat{x}$ to assess whether they
are naturally good looking \cite{Heusel2017}. However, since it is
hard to say whether a synthetic tabular sample looks real or fake
with bare eyes, evaluating a tabular generation method often follows
the ``train on synthetic, test on real (TSTR)'' approach \cite{Esteban2017,Jordon2019}.
It is a common practice to generate synthetic samples $\hat{x}$ and
their labels $\hat{y}$, then use them to train ML predictive models
and compute performance scores on a real test set. A better score
means a better tabular generation method. Figure \ref{fig:Training-and-evaluation}
illustrates the training and evaluation phases.

Among existing approaches, generative adversarial network (GAN)-based
methods are widely used for tabular data generation \cite{Choi2017,Srivastava2017,Park2018,Xu2019,Ashrapov2020,Kim2021}.
Recently, large language models (LLM)-based methods are proposed,
and they show promising results \cite{Borisov2023,zhang2023generative}.
Compared to GAN-based methods, LLM-based methods have some advantages.
First, since they do not require heavy data pre-processing steps e.g.
encoding categorical data or normalizing continuous data, they avoid
information loss and artificial introduction. Second, since they represent
tabular data as text instead of numerical, they can capture the context
knowledge among variables e.g. the relationship between \textit{Age}
and \textit{Marriage}.

\begin{figure*}
\begin{centering}
\includegraphics[scale=0.55]{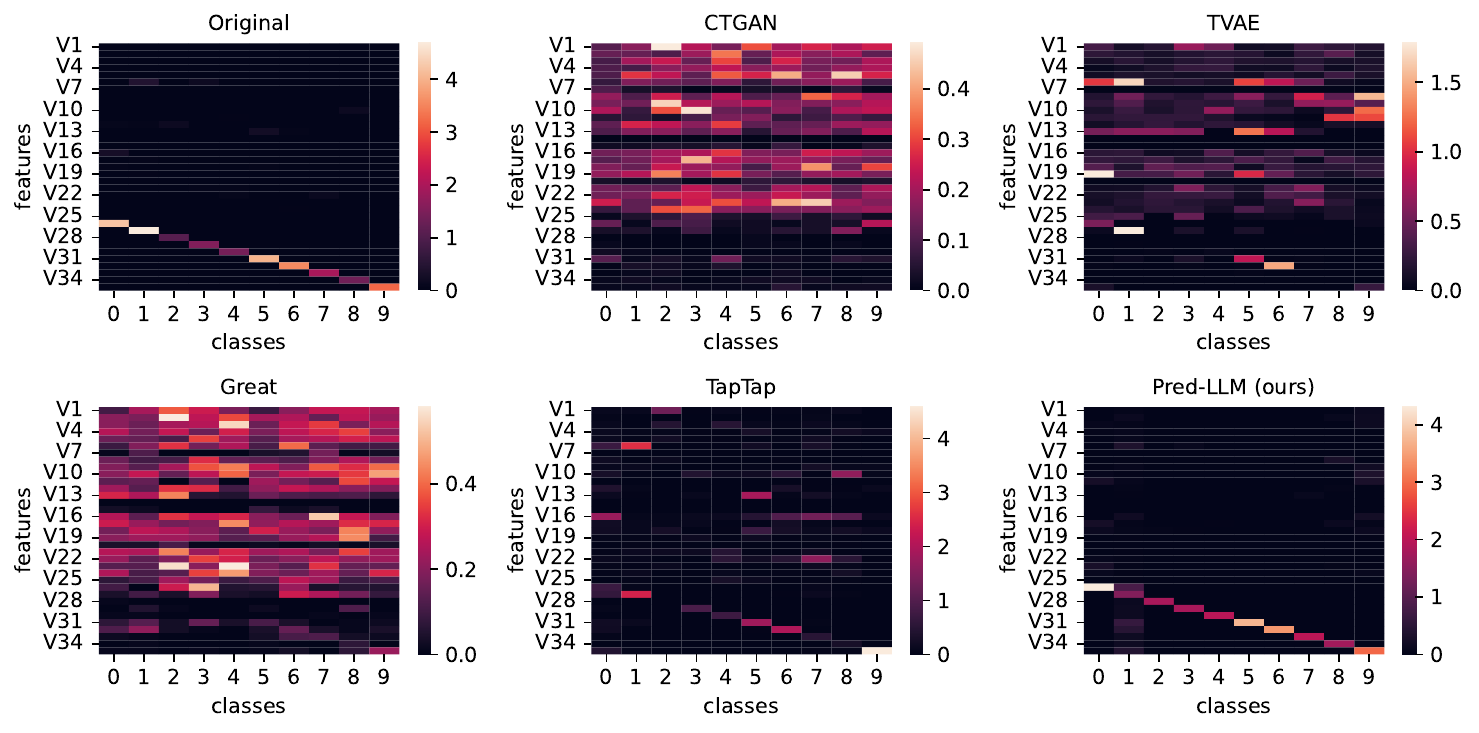}
\par\end{centering}
\caption{\label{fig:The-ground-truth-feature-label}The ground-truth feature-class
correlation on the original dataset \textit{Cardiotocography} (1\protect\textsuperscript{st}
plot). Existing methods cannot capture this correlation on the synthetic
data. Only our method \textbf{Pred-LLM} can capture it.}

\end{figure*}

To generate a sample $\hat{x}$ that has $M$ features $\{X_{1},...,X_{M}\}$
and a label $\hat{y}$, most existing methods treat the target variable
$Y$ as a regular feature $X_{M+1}$ \cite{Srivastava2017,Xu2019,Kim2021,Borisov2023}.
However, these methods may not correctly capture the correlation between
$X$ and $Y$, which is very important for training predictive models
in the evaluation phase (see Figure \ref{fig:Training-and-evaluation}).
Some methods generate only $\hat{x}$, and use an \textit{external
classifier} trained on the real dataset to predict $\hat{y}$ \cite{zhang2023generative}.
However, this method is cumbersome since it needs two standalone models
-- a generative model and a predictive model. As shown in Figure
\ref{fig:The-ground-truth-feature-label}, we compute the importance
level of each feature for each class on the dataset \textit{Cardiotocography}
using the Shapley values \cite{lundberg2017unified}. We compare the
feature importance on the original data with the feature importance
on the synthetic data generated by various methods. Existing state-of-the-art
methods such as TVAE \cite{Xu2019}, CTGAN \cite{Xu2019}, Great \cite{Borisov2023},
and TapTap \cite{zhang2023generative} \textit{cannot capture the
ground-truth feature-class correlation on the real dataset}. This
suggests that the data generated with these methods does not closely
mimic the real data.

To address the problem, we propose a LLM-based method with three improvements.
First, we observe that existing LLM-based methods \textit{permute
both the features and the target variable} in the fine-tuning phase
to support \textit{arbitrary conditioning} i.e. the capacity to generate
data conditioned on an arbitrary set of variables. However, this way
has a significant disadvantage. In the attention matrix of LLM, each
token only pays attention to every token before it, but not any after
it. As a result, if the target variable is shuffled to the beginning
positions and most of the features are after it in the sequences,
then there are no attention links from the features to the target
variable. Therefore, permuting both the features and the target variable
may cause the LLM to fail in capturing the correlation between $X$
and $Y$. We propose a simple yet effective trick to fix this issue.
For each tabular sample, we\textit{ only permute the features while
keeping the target variable at the end}. Our permutation strategy
has two reasons. First, we shuffle the features to enable arbitrary
conditioning. Second, we fix the target variable at the end to ensure
the LLM can attend it through all features. Our permutation strategy
helps our method to correctly learn the influences of the features
on the target variable as shown in Figure \ref{fig:The-ground-truth-feature-label}.

Second, in the sampling phase (i.e. the generation phase of synthetic
tabular samples), current LLM-based methods follow the \textit{class-conditional
sampling} style. In particular, they sample a label from the distribution
of the target variable and use it as the initial token to generate
the remaining tokens. However, this way may not be suitable for our
LLM since we keep the target variable at the end of the token list.
Given the target variable as the first token, our LLM may generate
the next tokens poorly as the attention links from the target variable
to the features are unavailable. Thus, we propose to use\textit{ each
feature as a condition}. In particular, we uniformly sample a feature
from the feature list and sample a value from the feature's distribution
to be the initial token for the generation process. We call our strategy
\textit{feature-conditional sampling}.

Finally, after generating samples $\hat{x}$, we construct prompts
based on $\hat{x}$ to query their labels $\hat{y}$. We explicitly
generate $\hat{y}$ from the conditional probability $p(\hat{y}\mid\hat{x})$
instead of simultaneously generating $\hat{x}$ and $\hat{y}$ from
the joint distribution $p(\hat{x},\hat{y})$ \cite{Xu2019}. This
mechanism helps us to better generate the corresponding label $\hat{y}$
for each synthetic sample $\hat{x}$. It also avoids using an external
classifier to predict $\hat{y}$ from $\hat{x}$ \cite{zhang2023generative},
which may be cumbersome in some cases and may not capture the feature
output contexts. Further, our experiments show that when the distributions
of the real dataset and the synthetic dataset are not similar, the
external classifier mostly predicts wrong labels for the synthetic
samples.

To summarize, we make the following contributions.

(1) We propose a LLM-based method (called \textbf{Pred-LLM}) for tabular
data generation. With three improvements in fine-tuning, sampling,
and label-querying phases, our method can generate realistic samples
that better capture the correlation between the features and the target
variable in the real data.

(2) We extensively evaluate our method \textbf{Pred-LLM} on 20 tabular
datasets and compare it with 10 SOTA baselines. \textbf{Pred-LLM}
is significantly better than other methods in most cases.

(3) In addition to showing the great benefits of our synthetic samples
in downstream predictive tasks, we also analyze other metrics to measure
their quality and diversity to show our method's superiority over
other methods.

\section{Related Work\label{sec:Related-Work}}

\subsection{Machine learning for tabular data}

Since ML methods have shown lots of successes on other data types
e.g. image, text, graph, they are recently being applied to tabular
data. These works can be categorized into five groups. \textit{Table
question answering} provides the answer for a given question over
tables \cite{yin-etal-2020-tabert,herzig-etal-2020-tapas}. \textit{Table
fact verification} assesses if an assumption is true for a given table
\cite{chen2019tabfact,zhang2020table}. \textit{Table to text} describes
a given table in text \cite{bao2018table,andrejczuk-etal-2022-table}.
\textit{Table structure understanding} aims to identify the table
characteristics e.g. column types, relations \cite{tang2021rpt,sui2024table}.
\textit{Table prediction} is the most popular task that predicts a
label for the target variable (e.g. \textit{Income}) based on a set
of features (e.g. \textit{Age} and \textit{Education}). Different
from other tasks where deep learning methods are dominant, traditional
ML methods like random forest, LightGBM \cite{ke2017lightgbm}, and
XGBoost \cite{chen2016xgboost} still outperform deep learning counterparts
in tabular prediction tasks.

\subsection{Generative models for tabular data}

Inspired by lots of successes of generative models in image data generation,
researchers have explored different ways to adapt them to tabular
data generation. Most tabular generation methods are based on GAN
models \cite{Goodfellow2014}, including \textit{MedGAN} \cite{Choi2017},
\textit{VeeGAN} \cite{Srivastava2017}, \textit{TableGAN} \cite{Park2018},
\textit{CTGAN} \cite{Xu2019}, \textit{CopulaGAN} \cite{Patki2016},
\textit{TabGAN} \cite{Ashrapov2020}, and \textit{OCTGAN} \cite{Kim2021}.
Some methods are based on other generative models e.g. \textit{TVAE}
\cite{Xu2019} uses Variational Autoencoder (VAE) \cite{Kingma2019},
\textit{Great} \cite{Borisov2023} and \textit{TapTap} \cite{zhang2023generative}
use Large Language Models (LLM) \cite{Beltagy2019}.

CTGAN \cite{Xu2019} is one of the most common methods, which has
three contributions to improve the modeling process. First, it applies
different activation functions to generate mixed data types. Second,
it normalizes a continuous value corresponding to its mode-specific.
Finally, it uses a conditional generator to address the data imbalance
problem in categorical columns. Although these proposals greatly improve
the quality of the synthetic data, most of them relate to the \textit{pre-processing}
tasks while the core training process of CTGAN is still based on the
Wasserstein GAN (WGAN) loss \cite{Arjovsky2017}. Many methods were
extended from CTGAN e.g. TabGAN \cite{Ashrapov2020} and OCTGAN \cite{Kim2021}.

\subsection{Large language models for tabular data}

The biggest weakness of GAN-based methods is that they require \textit{heavy
pre-processing} steps to model data efficiently. These steps may cause
the loss of important information or the introduction of artifacts.
For example, when the categorical variables are encoded into a one-hot
encoding (i.e. a numeric vector), it implies an artificial ordering
among the values \cite{borisov2022deep}. To overcome these problems,
LLM-based methods are recently proposed for tabular data generation
\cite{Borisov2023,zhang2023generative}. Compared to other methods,
LLM-based methods offer three great benefits: (1) preventing information
loss and artificial characteristics, (2) capturing problem specific
context, and (3) supporting arbitrary conditioning.

Existing LLM-based methods can generate realistic tabular data that
have various applications such as data privacy \cite{gulati2024tabmt},
data augmentation \cite{seedat2024curated}, class imbalance \cite{yang2024language},
and few-shot classification \cite{hegselmann2023tabllm}. However,
as shown in Figure \ref{fig:The-ground-truth-feature-label}, they
may not capture the correlation between the features $X$ and the
target variable $Y$. \textit{We are the first to propose a LLM-based
method for tabular data generation, which accurately captures the
correlation between $X$ and $Y$}.

\section{Framework\label{sec:Framework}}

\subsection{Problem definition}

Given a \textit{real} tabular dataset ${\cal D}_{real}=\{x_{i},y_{i}\}_{i=1}^{N}$,
each row is a pair of a sample $x_{i}$ with $M$ features $\{X_{1},...,X_{M}\}$
and a label $y_{i}$ (a value of the target variable $Y$). Our goal
is to learn a data synthesizer $G$ from ${\cal D}_{real}$, and then
use $G$ to generate a \textit{synthetic} tabular dataset ${\cal D}_{fake}=\{\hat{x}_{i},\hat{y}_{i}\}_{i=1}^{N}$.
Following other works \cite{Xu2019,Kim2021,Borisov2023}, we evaluate
the quality of ${\cal D}_{fake}$ by measuring the accuracy or MSE
of a ML predictive model trained on ${\cal D}_{fake}$ and tested
on a held-out test dataset ${\cal D}_{test}$ (shown in Figure \ref{fig:Training-and-evaluation}).
A better score means a better ${\cal D}_{fake}$.

\subsection{The proposed method}

\begin{figure*}
\begin{centering}
\includegraphics[scale=0.39]{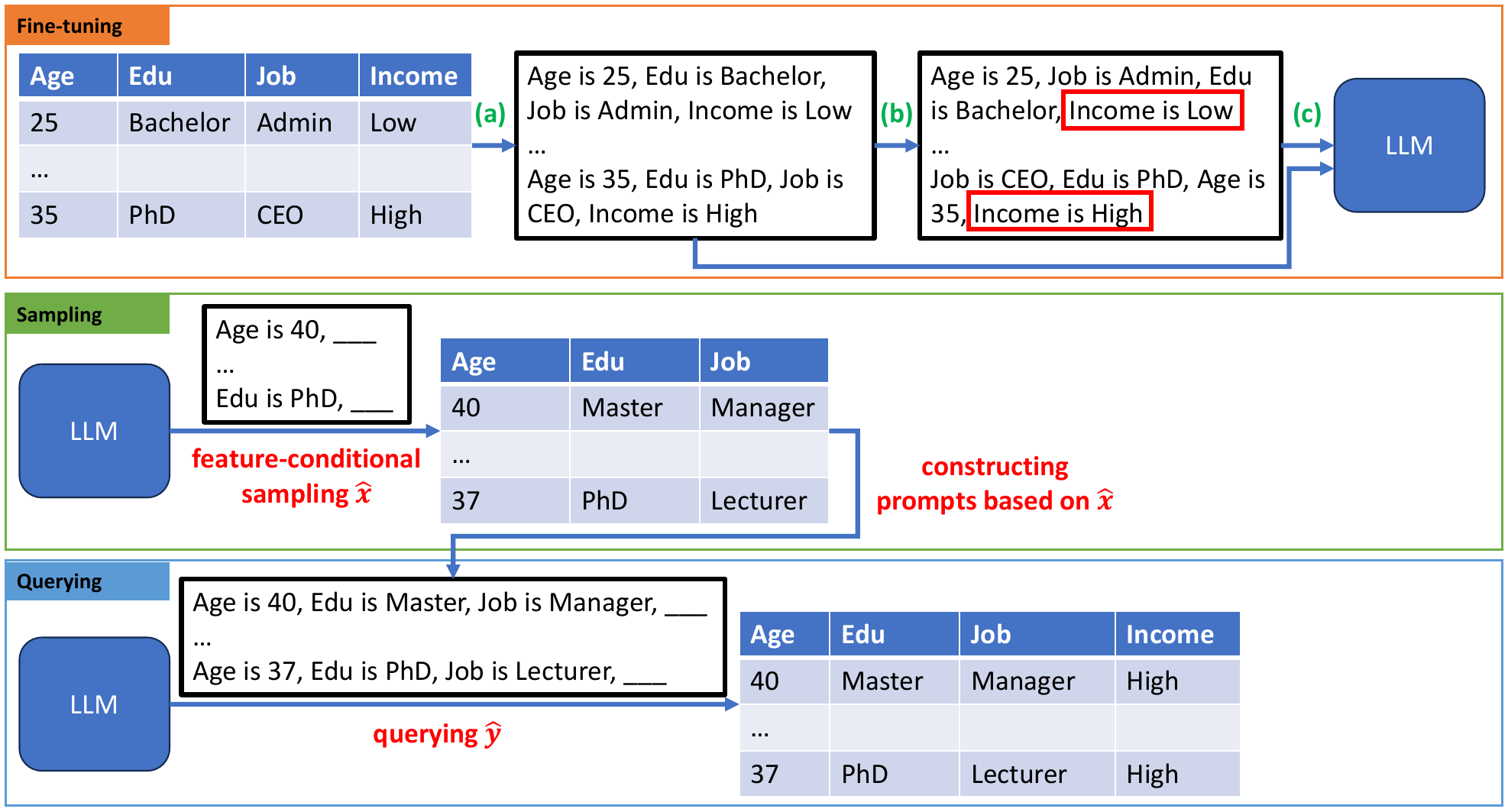}
\par\end{centering}
\caption{\label{fig:Overview-of-Pred-LLM}Our method Pred-LLM includes three
phases: (1) fine-tuning a pre-trained LLM with the real dataset, (2)
sampling synthetic samples conditioned on each feature, and (3) constructing
prompts based on the generated data to query labels.}
\end{figure*}

We propose a LLM-based method (called \textbf{Pred-LLM}) to generate
a synthetic dataset that mimics the real dataset and captures the
correlation between $X$ and $Y$ accurately. As shown in Figure \ref{fig:Overview-of-Pred-LLM},
our method has three phases: (1) fine-tuning a pre-trained LLM with
tabular data, (2) generating samples $\hat{x}$ conditioned on each
feature $X_{i}$, and (3) constructing prompts based on $\hat{x}$
to query labels $\hat{y}$.

\subsubsection{Fine-tuning}

In the first phase \textit{fine-tuning} (top plot), we have three
steps: (a) \textit{textual encoding}, (b) \textit{permutation}, and
(c) \textit{fine-tuning a pre-trained LLM}.

\textbf{(a) Textual encoding.} As our method uses LLM to generate
tabular data, following other works \cite{Borisov2023,zhang2023generative}
we convert each sample $x_{i}$ and its label $y_{i}$ into a \textit{sentence}.
Although there are several methods to serialize a tabular row \cite{hegselmann2023tabllm},
we use a simple strategy that transforms the $i$\textsuperscript{th}
row $r_{i}=[X_{1}=v_{i,1},...,X_{M}=v_{i,M},Y=y_{i}]$ into a corresponding
sentence $s_{i}=\text{"}X_{1}\text{ is }v_{i,1},...,X_{M}\text{ is }v_{i,M},Y\text{ is }y_{i}\text{"}$,
where $\{X_{1},...,X_{M}\}$ are feature names, $v_{i,j}$ is the
value of the $j$\textsuperscript{th} feature of the $i$\textsuperscript{th}
sample, $Y$ is the target variable and $y_{i}$ is its value. For
example, the first row $[Age=25,Edu=Bachelor,Job=Admin,Income=Low]$
is converted into ``\textit{Age is 25, Edu is Bachelor, Job is Admin,
Income is Low}''.

\textbf{(b) Permutation.} Recall that existing LLM-based methods permute
both the features $X$ and the target variable $Y$. We call this
strategy \textit{permute\_xy}. Formally, given a sentence $s_{i}=\text{"}X_{1}\text{ is }v_{i,1},...,X_{M}\text{ is }v_{i,M},Y\text{ is }y_{i}\text{"}$,
we re-write it in a short form $s_{i}=[a_{i,1},...,a_{i,M},a_{i,M+1}]$,
where $a_{i,j}=\text{"}X_{j}\text{ is }v_{i,j}\text{"}$ with $j\in\{1,...,M\}$,
$a_{i,M+1}=\text{"}Y\text{ is }y_{i}\text{"}$, and $[\cdot]$ denotes
the concatenation operator. The permutation strategy \textit{permute\_xy}
applies a \textit{permutation function} $P$ to randomly shuffle the
order of the features and the target variable. This step results in
a permuted sentence $s_{i}=[a_{i,k_{1}},...,a_{i,k_{M}},a_{i,k_{M+1}}]$,
where $[k_{1},...,k_{M},k_{M+1}]=P([1,...,M,M+1])$. For example,
the sentence ``\textit{Age is 25, Edu is Bachelor, Job is Admin,
Income is Low}'' is permuted to ``\textit{Income is Low, Edu is
Bachelor, Job is Admin, Age is 25}''.

In contrast, we only permute the features $X$ while fixing the target
variable $Y$ at the end. We call our strategy \textit{permute\_x}.
Formally, given a sentence $s_{i}=[a_{i,1},...,a_{i,M},a_{i,M+1}]$,
we apply the permutation function $P$ to $M$ features only, which
results in a permuted sentence $s_{i}=[a_{i,k_{1}},...,a_{i,k_{M}},a_{i,M+1}]$,
where $[k_{1},...,k_{M}]=P([1,...,M])$. For example, the sentence
``\textit{Age is 25, Edu is Bachelor, Job is Admin, Income is Low}''
is permuted to ``\textit{Age is 25, Job is Admin, Edu is Bachelor,
Income is Low}''. Note that the target variable ``Income'' and
its value ``Low'' are at the end of the permuted sentence.

\begin{figure}[th]
\begin{centering}
\includegraphics[scale=0.3]{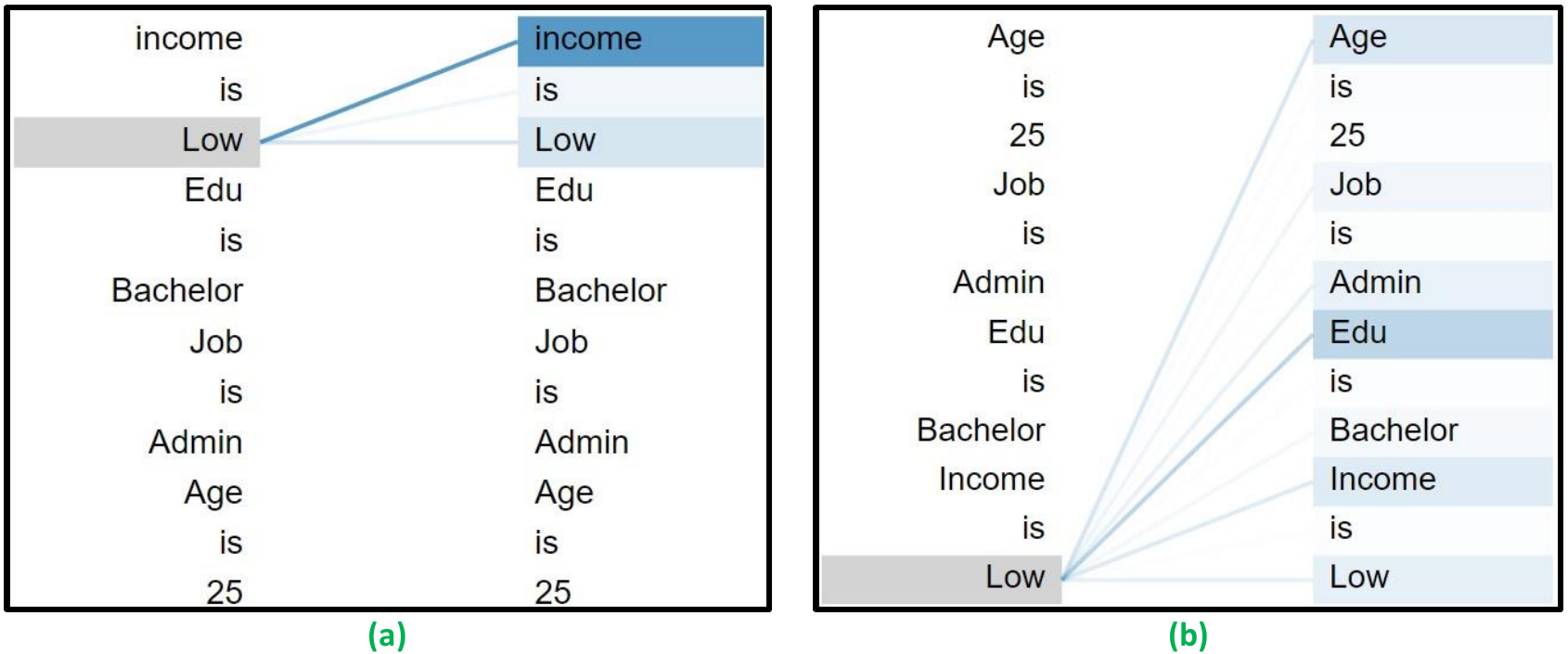}
\par\end{centering}
\caption{\label{fig:Attention-visualization}Attention visualization. Existing
LLM-based methods permute both $X$ and $Y$. We call this strategy
\textit{permute\_xy}. When the target variable ``Income'' is shuffled
to the beginning of the sentence, there is no attention links between
the features and the label ``Low'', as shown in (a). In contrast,
we only permute $X$ while fixing $Y$ at the end of the sentence.
We call our strategy \textit{permute\_x}. The attention mechanism
of our LLM can learn some correlations between the label ``Low''
and other features ``Age'' and ``Edu'', as shown in (b).}
\end{figure}

Our permutation strategy ensures that our LLM can learn the attention
links from the features $X$ to the target variable $Y$, which helps
to capture the correlation between $X$ and $Y$. In Figure \ref{fig:Attention-visualization},
we use BertViz \cite{vig-2019-multiscale} to visualize the attention
links learned by ChatGPT-2 -- the LLM model used in LLM-based methods
such as Great \cite{Borisov2023}, TapTap \cite{zhang2023generative},
and our Pred-LLM. As Great and TapTap use the permutation strategy
\textit{permute\_xy}, there is no attention link between the label
``Low'' with other features when the target variable ``Income''
is shuffled to the beginning of the sentence, as shown in (a). As
our method uses the permutation strategy \textit{permute\_x}, our
attention matrix can capture some strong correlations between the
label ``Low'' and the features ``Age'' and ``Edu'', as shown
in (b).

\textbf{(c) Fine-tuning.} While other LLM-based methods use the \textit{permute\_xy
}version of the dataset to fine-tune the LLM, we fine-tune our LLM
with the dataset permuted with our strategy \textit{permute\_x}. As
we also want to \textit{keep the original order of the features},
we augment the permuted data with the original data when fine-tuning
the model, as shown in step (c). This step is necessary because in
a later step where we construct a prompt based on a generated sample
$\hat{x}$ to query its label $\hat{y}$, we follow the original order
of the features in the real dataset.

We fine-tune our LLM following an auto-regressive manner i.e. we train
it to predict the next token given a set of observed tokens in each
encoded sentence. Given a textually encoded tabular dataset ${\cal S}=\{s_{i}\}_{i=1}^{N}$,
for each sentence $s_{i}\in{\cal S}$, we tokenize it into a sequence
of tokens $(c_{1},...,c_{l})=\text{tokenize}(s_{i})$, where $\{c_{1},...c_{l}\}$
are required tokens to describe the sentence $s_{i}$. Following an
auto-regressive manner, we factorize the probability of a sentence
$s_{i}$ into a product of output probabilities conditioned on previously
observed tokens:
\begin{equation}
p(s_{i})=p(c_{1},...,c_{l})=\prod_{k=1}^{l}p(c_{k}\mid c_{1},...,c_{k-1})\label{eq:fine-tune}
\end{equation}

We train the model to maximize the probability $\prod_{s_{i}\in{\cal S}}p(s_{i})$
of the entire training dataset ${\cal S}$. With this fine-tuning
process, we can use any pre-trained auto-regressive LLM e.g. GPT \cite{radford2018improving},
Transformer-XL \cite{dai2019transformer}, and Reformer \cite{kitaev2020reformer}.

\subsubsection{Sampling $\hat{x}$}

In the second phase \textit{sampling} (middle plot), we generate synthetic
samples $\hat{x}$. Given an input sequence of tokens $(c_{1},...,c_{k-1})$,
our fine-tuned LLM $Q$ returns logits over all possible follow-up
tokens:
\[
z=Q(c_{1},...,c_{k-1})
\]

The next token $c_{k}$ is sampled from a conditional probability
distribution defined by a \textit{softmax} function that is modified
by a temperature $T>0$ to soften the output distribution:
\begin{equation}
p(c_{k}\mid c_{1},...,c_{k-1})=\frac{e^{(z_{c_{k}}/T)}}{\sum_{c'\in{\cal C}}e^{(z_{c'}/T)}},\label{eq:sampling}
\end{equation}
where $e()$ is an exponential function and ${\cal C}$ is the complete
set of all unique tokens i.e. the vocabulary.

To generate $\hat{x}$, we need to initialize the sequence of tokens
$(c_{1},...,c_{k-1})$ and use it as a \textit{condition} for the
sampling step. As we employ permutation during the fine-tuning phase,
our method also supports arbitrary conditioning like other LLM-based
methods. In other words, we can use any set of features as a condition.
As mentioned in the literature, there are several ways to create a
condition. The most popular way is to use a pair of the target variable
and its value as a condition \cite{Borisov2023,zhang2023generative}.
In particular, a label $y_{i}$ is first sampled from the distribution
of the target variable $Y$ i.e. $y_{i}\sim p(Y)$. The condition
is then constructed as ``$Y\text{ is }y_{i}$'' e.g. ``Income is
Low''. We call this strategy \textit{class-conditional sampling}.

In contrast, our method follows a \textit{feature-conditional sampling}
style. We first uniformly sample a feature $X_{i}$ from the list
of features $\{X_{1},...,X_{M}\}$. We then sample a value from the
distribution of $X_{i}$ i.e. $v_{i}\sim p(X_{i})$. Finally, we construct
our condition as ``$X_{i}\text{ is }v_{i}$'' e.g. ``Age is 40''.

Our sampling strategy has an advantage. As we generate the features
$X$ before the target variable $Y$, our LLM can use all learned
attention links from $X$ to $Y$ to generate $Y$ more accurately.
However, it also has a weakness. As we cannot control when $Y$ is
generated, the correlations between $Y$ and some features may be
missed. For example, assume that we use $X_{1}$ as a condition. From
$X_{1}$, using Equation (\ref{eq:sampling}), we may generate $X_{3}$.
Conditioned on $(X_{1},X_{3})$, we may generate $Y$. In this case,
we generate $Y$ with only attentions coming from two features $X_{1}$
and $X_{3}$, and we may miss the attentions from the other features.
We address this problem in the next step.

\subsubsection{Querying $\hat{y}$}

In the final phase \textit{querying} (last plot), we construct prompts
to query our LLM for labels. After generating a synthetic sample $\hat{x}_{i}=[X_{1}=v_{i,1},...,X_{M}=v_{i,M}]$
with all features, we construct a prompt based on $\hat{x}_{i}$ as
``$X_{1}\text{ is }v_{i,1},...,X_{M}\text{ is }v_{i,M}$'' and use
it as a condition to sampling $\hat{y}_{i}$. Using this way, our
LLM can utilize the learned attention links from \textit{all features}
to better generate the target variable $Y$. Moreover, our LLM behaves
like a predictive model, and we generate $\hat{y}$ from a conditional
distribution $p(\hat{y}\mid\hat{x})$. This strategy works very well
thanks to the accurate correlation between $X$ and $Y$ captured
by our LLM.

\subsubsection{Algorithm}

Algorithm \ref{alg:Proposed-method-Pred-LLM} presents the pseudo-code
of our method Pred-LLM. It has three phases: fine-tuning, sampling,
and querying. In the fine-tuning phase (lines 4-8), we textually encode
each row to a sentence, permute it using our permutation strategy
\textit{permute\_x}, and create a new dataset ${\cal D}_{real}'$
that contains \textit{permuted} sentences and\textit{ original} sentences
(line 7). Finally, we use ${\cal D}_{real}'$ to fine-tune a pre-trained
LLM $Q$.

In the sampling phase (lines 17-22), we generate synthetic samples
$\hat{x}$ using our \textit{feature-conditional sampling} approach.
For each feature $X_{i}$, we use it as a condition to sample data
from our fine-tuned LLM $Q$. Note that the number of synthetic samples
conditioned on each $X_{i}$ is sampled equally (line 16).

In the querying phase (lines 24-27), we construct prompts based on
generated samples $\hat{x}$ to query their labels $\hat{y}$. In
this case, we use our LLM as a predictive model to generate labels.

\begin{algorithm}
\caption{\label{alg:Proposed-method-Pred-LLM}The Pred-LLM algorithm.}

\LinesNumbered

\KwIn{Real dataset ${\cal D}_{real}=\{x_{i},y_{i}\}_{i=1}^{N}$}

\KwOut{Fake dataset ${\cal D}_{fake}=\{\hat{x}_{i},\hat{y}_{i}\}_{i=1}^{N}$}

\Begin{

\textbf{Fine-tuning phase:}

${\cal D}_{real}'=\emptyset$;

\For{each row $r_{i}=(x_{i},y_{i})\in{\cal D}_{real}$ }{

encode $r_{i}$ to a sentence $s_{i}$;

permute $s_{i}$ to $s_{i}'=\text{permute\_x}(s_{i})$;

${\cal D}_{real}'={\cal D}_{real}'\cup\{s_{i}'\}\cup\{s_{i}\}$;

}

\For{each sentence $s_{i}\in{\cal D}_{real}'$ }{

$(c_{1},...,c_{l})=\text{tokenize}(s_{i})$;

fine-tune a pre-trained LLM $Q$ with $(c_{1},...,c_{l})$ using Eq.
(\ref{eq:fine-tune});

}

\textbf{Sampling phase:}

${\cal D}_{fake}=\emptyset$;

recall that each sample $x_{i}$ has $M$ features $\{X_{1},...,X_{M}\}$;

compute no. of generated samples conditioned on each feature $X_{i}$:
$n=\nicefrac{N}{M}$;

\For{each feature $X_{i}\in\{X_{1},...,X_{M}\}$ }{

sample a value $v_{i}\sim p(X_{i})$;

construct condition ``$X_{i}\text{ is }v_{i}$'';

sample $n$ synthetic samples $\{\hat{x}_{i}\}_{i=1}^{n}=Q(\text{\textquotedblleft}X_{i}\text{ is }v_{i}\text{\textquotedblright};n)$
using Eq. (\ref{eq:sampling});

${\cal D}_{fake}={\cal D}_{fake}\cup\{\hat{x}_{i}\}_{i=1}^{n}$;

}

\textbf{Querying phase:}

\For{each synthetic sample $\hat{x}_{i}\in{\cal D}_{fake}$ }{

construct prompt ``$X_{1}\text{ is }v_{i,1},...,X_{M}\text{ is }v_{i,M}$'';

query $Q$ for $\hat{y}_{i}$ using Eq. (\ref{eq:sampling}): $\hat{y}_{i}=Q(\text{\textquotedblleft}X_{1}\text{ is }v_{i,1},...,X_{M}\text{ is }v_{i,M}\text{\textquotedblright};1)$;

}

return ${\cal D}_{fake}=\{\hat{x}_{i},\hat{y}_{i}\}_{i=1}^{N}$;

}
\end{algorithm}

\section{Experiments\label{sec:Experiments}}

We conduct extensive experiments to show that our method is much better
than other methods under different qualitative and quantitative metrics.

\subsection{Experiment settings}

\subsubsection{Datasets}

We evaluate our method on 20 real-world tabular datasets. They are
commonly used in predictive and generation tasks \cite{Xu2019,Nguyen2020,Kim2021,Borisov2023,zhang2023generative}.
The details of these datasets are provided in Table \ref{tab:Characteristics-datasets}.

\begin{table}
\caption{\label{tab:Characteristics-datasets}Characteristics of 20 benchmark
datasets. $N/M/K$ indicates the number of samples/features/classes.
\textit{Due to space limitation, we shorten some long-name datasets
in parenthesis.}}

\centering{}%
\begin{tabular}{|l|r|r|r|l|}
\hline 
\rowcolor{header_color}Dataset & $N$ & $M$ & $K$ & Source\tabularnewline
\hline 
\hline 
\textit{iris} & 150 & 4 & 3 & scikit-learn\tabularnewline
\hline 
\rowcolor{even_color}\textit{breast\_cancer} (breast) & 569 & 30 & 2 & scikit-learn\tabularnewline
\hline 
\textit{blood\_transfusion} (blood) & 748 & 4 & 2 & OpenML\tabularnewline
\hline 
\rowcolor{even_color}\textit{steel\_plates\_fault} (steel) & 1,941 & 33 & 2 & OpenML\tabularnewline
\hline 
\textit{diabetes} & 768 & 8 & 2 & OpenML\tabularnewline
\hline 
\rowcolor{even_color}\textit{australian} & 690 & 14 & 2 & OpenML\tabularnewline
\hline 
\textit{balance\_scale} (balance) & 625 & 4 & 3 & OpenML\tabularnewline
\hline 
\rowcolor{even_color}\textit{compas} & 4,010 & 10 & 2 & \cite{Nguyen2021}\tabularnewline
\hline 
\textit{bank} & 4,521 & 14 & 2 & \cite{Nguyen2021}\tabularnewline
\hline 
\rowcolor{even_color}\textit{adult} & 30,162 & 13 & 2 & UCI\tabularnewline
\hline 
\textit{qsar\_biodeg} (qsar) & 1,055 & 41 & 2 & OpenML\tabularnewline
\hline 
\rowcolor{even_color}\textit{phoneme} & 5,404 & 5 & 2 & OpenML\tabularnewline
\hline 
\textit{waveform} & 5,000 & 40 & 3 & OpenML\tabularnewline
\hline 
\rowcolor{even_color}\textit{churn} & 5,000 & 20 & 2 & OpenML\tabularnewline
\hline 
\textit{kc1} & 2,109 & 21 & 2 & OpenML\tabularnewline
\hline 
\rowcolor{even_color}\textit{kc2} & 522 & 21 & 2 & OpenML\tabularnewline
\hline 
\textit{cardiotocography} (cardio) & 2,126 & 35 & 10 & OpenML\tabularnewline
\hline 
\rowcolor{even_color}\textit{abalone} & 4,177 & 8 & regression & Kaggle\tabularnewline
\hline 
\textit{fuel} & 639 & 8 & regression & Kaggle\tabularnewline
\hline 
\rowcolor{even_color}\textit{california} & 20,640 & 8 & regression & scikit-learn\tabularnewline
\hline 
\end{tabular}
\end{table}

\subsubsection{Evaluation metric\label{subsec:Evaluation-metric}}

To evaluate the performance of tabular generation methods, we use
the synthetic data in predictive tasks (Figure \ref{fig:Training-and-evaluation})
similar to other works \cite{Xu2019,Kim2021,Borisov2023}. For each
dataset, we randomly split it into 80\% for the real set ${\cal D}_{real}$
and 20\% for the test set ${\cal D}_{test}$. We train tabular generation
methods on ${\cal D}_{real}$ to generate the synthetic set ${\cal D}_{fake}$,
where $\mid{\cal D}_{fake}\mid=\mid{\cal D}_{real}\mid$. Finally,
we train XGBoost on ${\cal D}_{fake}$ and compute its accuracy/MSE
on ${\cal D}_{test}$. We repeat each method \textit{three times with
random seeds} and report the average score along with its standard
deviation.

We also compute other metrics between ${\cal D}_{real}$ and ${\cal D}_{fake}$
to measure the \textit{quality} and \textit{diversity} of the synthetic
samples:

(1) \textit{Discriminator score} \cite{Borisov2023}: We train a XGBoost
discriminator to differentiate the synthetic data from the real data.
The score of $0$ means two datasets are indistinguishable whereas
$1$ means two datasets are totally distinguishable (i.e. synthetic
samples are easily detectable). \textit{A lower score is better}.

(2) \textit{Inverse Kullback--Leibler divergence} \cite{qian2023synthcity}:
We compute the inverse of the Kullback-Leibler averaged over all features
to measure how the distribution of the synthetic data $p({\cal D}_{fake})$
is different from the distribution of the real data $p({\cal D}_{real})$.
Let $X_{i}$ be a feature (i.e. a column) in the real table ${\cal D}_{real}$
and $\hat{X}_{i}$ be its corresponding feature in the synthetic table
${\cal D}_{fake}$, the inverse KL score between $p({\cal D}_{real})$
and $p({\cal D}_{fake})$ is $InvKL=\frac{1}{M}\sum_{i=1}^{M}\frac{1}{1+\text{KL}(p(X_{i})\mid\mid p(\hat{X}_{i}))}$.
As $p(X_{i})>0$, $\text{KL}(p(X_{i})\mid\mid p(\hat{X}_{i}))$ is
finite, which results in $InvKL\in[0,1]$. The score of $0$ means
two datasets are from different distributions whereas $1$ means they
are from the same distribution. \textit{A higher score is better}.

(3) \textit{Density} \cite{naeem2020reliable}: We compute the density
score $Den=\frac{1}{kN}\sum_{j=1}^{N}\sum_{i=1}^{N}\mathbb{I}_{\hat{x}_{j}\in B(x_{i},\text{NND}_{k}(x_{i}))}$,
where $B(x_{i},\text{NND}_{k}(x_{i}))$ is the sphere around $x_{i}$
with the radius $\text{NND}_{k}(x_{i})$ being the Euclidean distance
from $x_{i}$ to its $k$-th nearest neighbor (we set $k=2$). It
measures how many synthetic samples $\hat{x}_{j}$ reside in the neighbor
of the real samples $x_{i}$. It measures the \textit{quality} of
the synthetic samples as it shows how the synthetic samples are similar/close
to the real samples. \textit{A higher score is better}.

(4) \textit{Coverage} \cite{naeem2020reliable}: We compute the coverage
score $Cov=\frac{1}{N}\sum_{i=1}^{N}\mathbb{I}_{\exists j\text{ s.t. }\hat{x}_{j}\in B(x_{i},\text{NND}_{k}(x_{i}))}$.
It measures the fraction of real samples $x_{i}$ whose neighborhoods
contain at least one synthetic sample $\hat{x}_{j}$. It measures
the \textit{diversity} of the synthetic samples as it shows how the
synthetic samples resemble the variability of the real samples. \textit{A
higher score is better}.

\subsubsection{Baselines}

We compare with 10 SOTA baselines. They include GAN-based methods
(\textit{CopulaGAN} \cite{Patki2016}\textit{, MedGAN} \cite{Choi2017},
\textit{VeeGAN} \cite{Srivastava2017}, \textit{TableGAN} \cite{Park2018},
\textit{CTGAN} \cite{Xu2019}, \textit{TabGAN} \cite{Ashrapov2020},
and \textit{OCTGAN} \cite{Kim2021}), VAE-based method (\textit{TVAE}
\cite{Xu2019}), and LLM-based methods (\textit{Great} \cite{Borisov2023}
and \textit{TapTap} \cite{zhang2023generative}).

The \textit{Original} method means the predictive model trained with
the real dataset ${\cal D}_{real}$. We use XGBoost for the predictive
model since it is one of the most popular models for tabular data
\cite{shwartz2022tabular}. Following \cite{Borisov2023,zhang2023generative},
we use the distilled version of ChatGPT-2 for our LLM, and train it
with \textit{batch\_size=32} and \textit{\#epochs=50}. We set $T=0.7$
for Equation (\ref{eq:sampling}).

\subsection{Results and discussions}

\subsubsection{Downstream predictive tasks}

Table \ref{tab:Accuracy/MSE} reports accuracy and MSE of each method
on 20 benchmark datasets. \textit{We recall that the procedure to
compute the scores is described in Figure \ref{fig:Training-and-evaluation}}.
Our method Pred-LLM performs better than other methods. Among 20 datasets,
Pred-LLM is the best performing on nine datasets and the second-best
on another nine datasets.

\begin{table*}
\caption{\label{tab:Accuracy/MSE}Accuracy/MSE $\pm$ (std) of each tabular
generation method on 20 datasets. \textbf{Bold} and \uline{underline}
indicate the best and second-best methods. ``\textit{Accuracy=0}''
means the method cannot generate a \textit{valid} ${\cal D}_{fake}$
to train predictive models (e.g. ${\cal D}_{fake}$ contains only
one class). \textit{Due to space limitation, we use ``{*}'' as an
abbreviation of ``GAN'' e.g. CT{*} means CTGAN}.}

\centering{}%
\begin{tabular}{|l|r|r|r|r|r|r|r|r|r|r|r|r|}
\hline 
\rowcolor{header_color}\textbf{Accuracy} & Original & Copula{*} & Med{*} & Vee{*} & Table{*} & CT{*} & Tab{*} & OCT{*} & TVAE & Great & TapTap & Pred-LLM\tabularnewline
\hline 
\hline 
iris & 0.9555 & 0.3445 & 0.0000 & 0.0000 & 0.0000 & 0.4444 & 0.2667 & 0.7222 & \uline{0.9000} & 0.3667 & 0.8778 & \textbf{0.9889}\tabularnewline
 & (0.031) & (0.057) & (0.000) & (0.000) & (0.000) & (0.150) & (0.072) & (0.103) & 0.055) & (0.098) & (0.069) & (0.016)\tabularnewline
\hline 
\rowcolor{even_color}breast & 0.9708 & 0.5409 & 0.4445 & 0.5439 & 0.0000 & 0.4796 & 0.3889 & 0.6667 & \uline{0.9123} & 0.5760 & 0.8889 & \textbf{0.9503}\tabularnewline
\rowcolor{even_color} & (0.004) & (0.062) & (0.132) & (0.124) & (0.000) & (0.092) & (0.039) & (0.186) & (0.031) & (0.097) & (0.017) & (0.015)\tabularnewline
\hline 
blood & 0.7444 & 0.7178 & 0.5733 & 0.5733 & 0.6667 & 0.7355 & \textbf{0.7645} & 0.6022 & 0.7445 & 0.6866 & 0.7422 & \uline{0.7489}\tabularnewline
 & (0.022) & (0.014) & (0.245) & (0.236) & (0.043) & (0.030) & (0.003) & (0.073) & (0.019) & (0.041) & (0.006) & (0.025)\tabularnewline
\hline 
\rowcolor{even_color}steel & 1.0000 & 0.5955 & 0.4139 & 0.4559 & 0.9280 & 0.6298 & 0.6564 & 0.6573 & 0.6829 & 0.6392 & \uline{0.9135} & \textbf{1.0000}\tabularnewline
\rowcolor{even_color} & (0.000) & (0.035) & (0.095) & (0.101) & (0.015) & (0.027) & (0.003) & (0.006) & (0.001) & (0.045) & (0.018) & (0.000)\tabularnewline
\hline 
diabetes & 0.7598 & 0.4827 & 0.5195 & 0.5325 & 0.6494 & 0.5909 & 0.6147 & 0.5757 & 0.7035 & 0.6039 & \textbf{0.7662} & \uline{0.7099}\tabularnewline
 & (0.037) & (0.046) & (0.102) & (0.090) & (0.058) & (0.060) & (0.021) & (0.089) & (0.030) & (0.014) & (0.024) & (0.045)\tabularnewline
\hline 
\rowcolor{even_color}australian & 0.8768 & 0.4299 & 0.5797 & 0.4541 & 0.8092 & 0.5459 & 0.5387 & 0.6015 & 0.7753 & 0.6836 & \textbf{0.8865} & \uline{0.8768}\tabularnewline
\rowcolor{even_color} & (0.006) & (0.015) & (0.095) & (0.028) & (0.036) & (0.106) & (0.018) & (0.048) & (0.033) & (0.036) & (0.017) & (0.026)\tabularnewline
\hline 
balance & 0.8613 & 0.4773 & 0.0000 & 0.1627 & 0.4373 & 0.4640 & 0.3947 & 0.4453 & 0.0000 & 0.4693 & \textbf{0.8613} & \uline{0.8507}\tabularnewline
 & (0.010) & (0.032) & (0.000) & (0.056) & (0.061) & (0.013) & (0.051) & (0.133) & (0.000) & (0.025) & (0.014) & (0.036)\tabularnewline
\hline 
\rowcolor{even_color}compas & 0.8317 & 0.8134 & 0.5208 & 0.5964 & 0.5719 & 0.8284 & \textbf{0.8379} & 0.6563 & 0.8209 & 0.8188 & 0.8313 & \uline{0.8317}\tabularnewline
\rowcolor{even_color} & (0.003) & (0.028) & (0.256) & (0.123) & (0.082) & (0.007) & (0.000) & (0.204) & (0.030) & (0.003) & (0.007) & (0.004)\tabularnewline
\hline 
bank & 0.8803 & 0.8656 & 0.6659 & 0.6302 & 0.7319 & 0.8556 & 0.8604 & 0.5901 & 0.8641 & 0.8685 & \uline{0.8840} & \textbf{0.8847}\tabularnewline
 & (0.007) & (0.025) & (0.155) & (0.258) & (0.126) & (0.028) & (0.015) & (0.417) & (0.008) & (0.003) & (0.002) & (0.002)\tabularnewline
\hline 
\rowcolor{even_color}adult & 0.8624 & 0.8284 & 0.5788 & 0.4367 & 0.7559 & 0.8261 & 0.7239 & 0.2498 & 0.8253 & 0.8460 & \uline{0.8528} & \textbf{0.8530}\tabularnewline
\rowcolor{even_color} & (0.000) & (0.008) & (0.231) & (0.081) & (0.019) & (0.004) & (0.020) & (0.353) & (0.005) & (0.002) & (0.003) & (0.001)\tabularnewline
\hline 
qsar & 0.8752 & 0.6477 & 0.3412 & 0.4771 & 0.6983 & 0.6493 & 0.6319 & 0.5972 & \uline{0.7899} & 0.5008 & 0.6619 & \textbf{0.7946}\tabularnewline
 & (0.012) & (0.020) & (0.008) & (0.132) & (0.041) & (0.026) & (0.038) & (0.071) & (0.020) & (0.037) & (0.023) & (0.024)\tabularnewline
\hline 
\rowcolor{even_color}phoneme & 0.8970 & 0.7018 & 0.5134 & 0.4983 & 0.7321 & 0.7333 & 0.6876 & 0.5183 & 0.7684 & 0.7700 & \textbf{0.8705} & \uline{0.8541}\tabularnewline
\rowcolor{even_color} & (0.006) & (0.006) & (0.111) & (0.061) & (0.021) & (0.028) & (0.029) & (0.164) & (0.009) & (0.025) & (0.001) & (0.007)\tabularnewline
\hline 
waveform & 0.8467 & 0.3817 & 0.0000 & 0.3297 & 0.8330 & 0.3867 & 0.3457 & 0.1147 & \uline{0.8350} & 0.3557 & 0.7473 & \textbf{0.8473}\tabularnewline
 & (0.004) & (0.028) & (0.000) & (0.002) & (0.002) & (0.032) & (0.014) & (0.162) & (0.005) & (0.023) & (0.020) & (0.005)\tabularnewline
\hline 
\rowcolor{even_color}churn & 0.9580 & 0.8450 & 0.6390 & 0.3987 & 0.8653 & 0.8423 & 0.8467 & 0.2793 & 0.8547 & 0.8333 & \textbf{0.9090} & \uline{0.8703}\tabularnewline
\rowcolor{even_color} & (0.002) & (0.012) & (0.047) & (0.308) & (0.003) & (0.012) & (0.002) & (0.301) & (0.021) & (0.006) & (0.006) & (0.017)\tabularnewline
\hline 
kc1 & 0.8547 & 0.7986 & \textbf{0.8468} & 0.6153 & 0.7093 & 0.7986 & 0.8460 & 0.5561 & \uline{0.8444} & 0.8160 & 0.8144 & 0.8389\tabularnewline
 & (0.006) & (0.036) & (0.006) & (0.325) & (0.034) & (0.041) & (0.000) & (0.393) & (0.002) & (0.013) & (0.006) & (0.005)\tabularnewline
\hline 
\rowcolor{even_color}kc2 & 0.8127 & 0.7619 & 0.4032 & 0.4095 & 0.0000 & 0.7714 & 0.7937 & 0.6540 & \textbf{0.8572} & 0.7841 & \uline{0.8254} & 0.8190\tabularnewline
\rowcolor{even_color} & (0.022) & (0.028) & (0.281) & (0.283) & (0.000) & (0.016) & (0.005) & (0.133) & (0.014) & (0.020) & (0.025) & (0.008)\tabularnewline
\hline 
cardio & 1.0000 & 0.2222 & 0.0000 & 0.0000 & 0.6229 & 0.2387 & 0.0000 & 0.0000 & 0.8052 & 0.1988 & \uline{0.8998} & \textbf{1.0000}\tabularnewline
 & (0.000) & (0.027) & (0.000) & (0.000) & (0.102) & (0.050) & (0.000) & (0.000) & (0.028) & (0.036) & (0.025) & (0.000)\tabularnewline
\hline 
\rowcolor{childheader_color}Average & 0.8816 & 0.6150 & 0.4141 & 0.4185 & 0.5889 & 0.6365 & 0.5999 & 0.4992 & 0.7637 & 0.6363 & \uline{0.8372} & \textbf{0.8658}\tabularnewline
\hline 
\hline\rowcolor{header_color}\textbf{MSE} & Original & Copula{*} & Med{*} & Vee{*} & Table{*} & CT{*} & Tab{*} & OCT{*} & TVAE & Great & TapTap & Pred-LLM\tabularnewline
\hline 
abalone & 5.2213 & 8.1660 & 43.6048 & 24.3119 & 6.1970 & 8.3744 & 40.1766 & 16.6444 & 7.4558 & 11.1428 & \textbf{5.4765} & \uline{5.6194}\tabularnewline
 & (0.305) & (0.576) & (11.052) & (3.997) & (0.056) & (0.403) & (15.986) & (2.962) & (0.405) & (1.154) & (0.382) & (0.470)\tabularnewline
\hline 
\rowcolor{even_color}fuel & 0.0760 & 15.9164 & 19.2074 & 26.2352 & 1.5805 & 22.5587 & 17.7842 & 20.7891 & 4.6480 & 15.9838 & \textbf{0.2237} & \uline{0.8260}\tabularnewline
\rowcolor{even_color} & (0.019) & (4.579) & (6.303) & (5.552) & (0.685) & (7.015) & (6.258) & (11.297) & (1.756) & (3.776) & (0.118) & (0.360)\tabularnewline
\hline 
california & 0.2696 & 0.7531 & 2.3306 & 2.8342 & 0.8680 & 0.7208 & 1.8239 & 1.4400 & 0.6248 & 0.3556 & \uline{0.3286} & \textbf{0.3058}\tabularnewline
 & (0.008) & (0.043) & (0.156) & (1.045) & (0.044) & (0.064) & (0.279) & (0.130) & (0.026) & (0.011) & (0.006) & (0.008)\tabularnewline
\hline 
\rowcolor{childheader_color}Average & 1.8556 & 8.2785 & 21.7143 & 17.7938 & 2.8818 & 10.5513 & 19.9282 & 12.9578 & 4.2429 & 9.1607 & \textbf{2.0096} & \uline{2.2504}\tabularnewline
\hline 
\end{tabular}
\end{table*}

For classification tasks, the average improvement of Pred-LLM over
TapTap (the runner-up baseline) is $\sim3\%$. CTGAN is the best GAN-based
method, followed by CopulaGAN. Other GAN-based baselines do not perform
as well. The LLM-based method Great is comparable with CTGAN. TVAE
often outperforms GAN-based methods since it has a reconstruction
loss to guarantee that the synthetic samples are similar to the real
samples. The same observation was also made in \cite{Xu2019}. For
regression tasks, Pred-LLM and TapTap perform comparably.

\subsubsection{Quality and diversity evaluation}

As described in Section \ref{subsec:Evaluation-metric}, we compute
four metrics. Discriminator and Density scores measure the \textit{quality
of synthetic samples}, Inverse KL score measures the \textit{quality
of the synthetic distribution}, and Coverage score measures the \textit{diversity
of synthetic samples}.

\begin{figure}[th]
\begin{centering}
\includegraphics[scale=0.43]{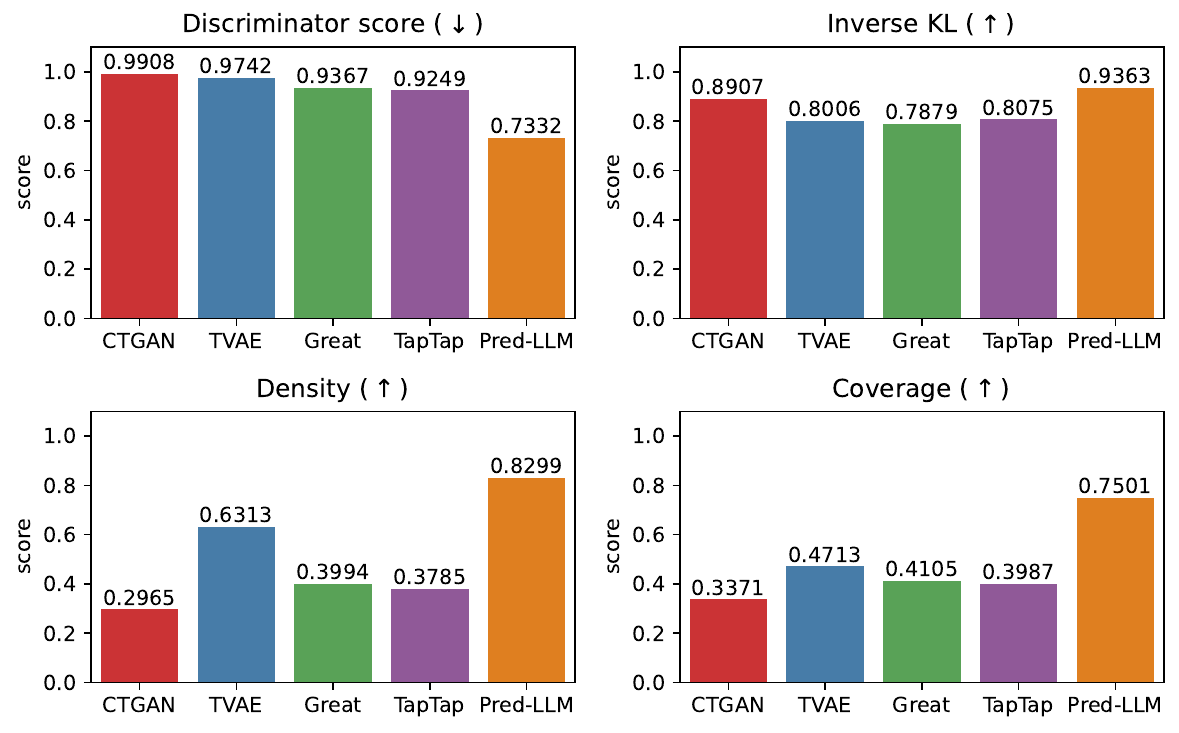}
\par\end{centering}
\caption{\label{fig:Quality-and-diversity}The average quality and diversity
scores over 20 datasets. $\downarrow$ means ``\textit{lower is better}''.
$\uparrow$ means ``\textit{higher is better}''.}
\end{figure}

Figure \ref{fig:Quality-and-diversity} presents the average scores
on top-5 methods on all datasets. Our method Pred-LLM significantly
outperforms other methods for all metrics. The Discriminator scores
show the baseline synthetic samples are easily detected by the discriminator
($>92\%$ chance) whereas only $73\%$ of our synthetic samples are
detectable. Another quality metric Density shows our synthetic samples
are much closer to the real samples than other synthetic samples (e.g.
$0.8299$ of Pred-LLM vs. $0.6313$ of TVAE). The Inverse KL shows
our synthetic samples and the real samples come from a similar distribution.
In terms of diversity, our synthetic samples are much more diverse
than others (e.g. $0.7501$ of Pred-LLM vs. $0.4713$ of TVAE).

High scores in both quality and diversity metrics show our synthetic
samples can capture the \textit{real data manifold}, which explains
their benefits in downstream tasks (see Table \ref{tab:Accuracy/MSE}).

\subsection{Ablation studies}

We analyze our method under different configurations.

\subsubsection{Effect investigation of various modifications}

We have three novel contributions in \textit{permutation}, \textit{sampling},
and \textit{label-querying} steps. Table \ref{tab:Effectiveness-of-different}
reports the accuracy for each modification compared to two LLM baselines
Great and TapTap. Recall that the baselines permute the features $X$
and the target variable $Y$ (denoted by \textit{permute\_xy}), and
they use $Y$ as a sampling condition (denoted by \textit{target\_variable}).
Great generates both $\hat{x}$ and $\hat{y}$ simultaneously (denoted
by \textit{querying $\hat{y}$=none}) whereas TapTap generates $\hat{x}$
first and then uses an external classifier to predict $\hat{y}$ (denoted
by \textit{querying $\hat{y}$=classifier}).

While Great only achieves $0.6363$, our modifications show improvements.
By using each feature as a sampling condition (denoted by \textit{each\_feature})
or prompting the LLM model for $\hat{y}$ (denoted by \textit{querying
$\hat{y}$=LLM}), we achieve up to $\sim0.73$ ($10\%$ improvement).
Combining these two proposals with fine-tuning our LLM on the dataset
permuted by our permutation strategy (denoted by\textit{ permute\_x}),
we achieve the best result at 0.8658 (last row). This study suggests
that each modification is useful, which greatly improves Great and
TapTap.

\begin{table}[th]
\caption{\label{tab:Effectiveness-of-different}Effectiveness of different
modifications in our Pred-LLM. The accuracy is averaged over 17 classification
datasets.}

\centering{}%
\begin{tabular}{|l|l|l|l|r|}
\hline 
\rowcolor{header_color} & Permutation & Sampling $\hat{x}$ & Querying $\hat{y}$ & \textbf{Accuracy}\tabularnewline
\hline 
\hline 
Great & \textit{permute\_xy} & \textit{target\_variable} & \textit{none} & 0.6363\tabularnewline
\hline 
TapTap & \textit{permute\_xy} & \textit{target\_variable} & \textit{classifier} & 0.8372\tabularnewline
\hline 
\multirow{6}{*}{Pred-LLM} & \textit{permute\_x} & \textit{target\_variable} & \textit{none} & 0.6191\tabularnewline
\cline{2-5} 
 & \textit{permute\_xy} & \textit{each\_feature} & \textit{none} & 0.7352\tabularnewline
\cline{2-5} 
 & \textit{permute\_xy} & \textit{target\_variable} & \textit{LLM} & 0.7307\tabularnewline
\cline{2-5} 
 & \textit{permute\_xy} & \textit{each\_feature} & \textit{LLM} & 0.7477\tabularnewline
\cline{2-5} 
 & \textit{permute\_x} & \textit{target\_variable} & \textit{LLM} & 0.8495\tabularnewline
\cline{2-5} 
 & \textbf{\textit{permute\_x}} & \textbf{\textit{each\_feature}} & \textbf{\textit{LLM}} & \textbf{0.8658}\tabularnewline
\hline 
\end{tabular}
\end{table}

\subsubsection{Distance to closest records (DCR) histogram}

Following \cite{Borisov2023}, we plot the DCR histogram to show \textit{our
synthetic dataset is} \textit{similar to the real dataset but} \textit{it
is not simply duplicated}. The DCR metric computes the distance from
a real sample to its closest neighbor in the synthetic dataset. Given
a real sample $x\in{\cal D}_{real}$, $\text{DCR}(x)=\min\{d(x,\hat{x}_{i})\mid\hat{x}_{i}\in{\cal D}_{fake}\}$.
We use the Euclidean distance for the distance function $d(\cdot)$.

\begin{figure}[H]
\begin{centering}
\includegraphics[scale=0.36]{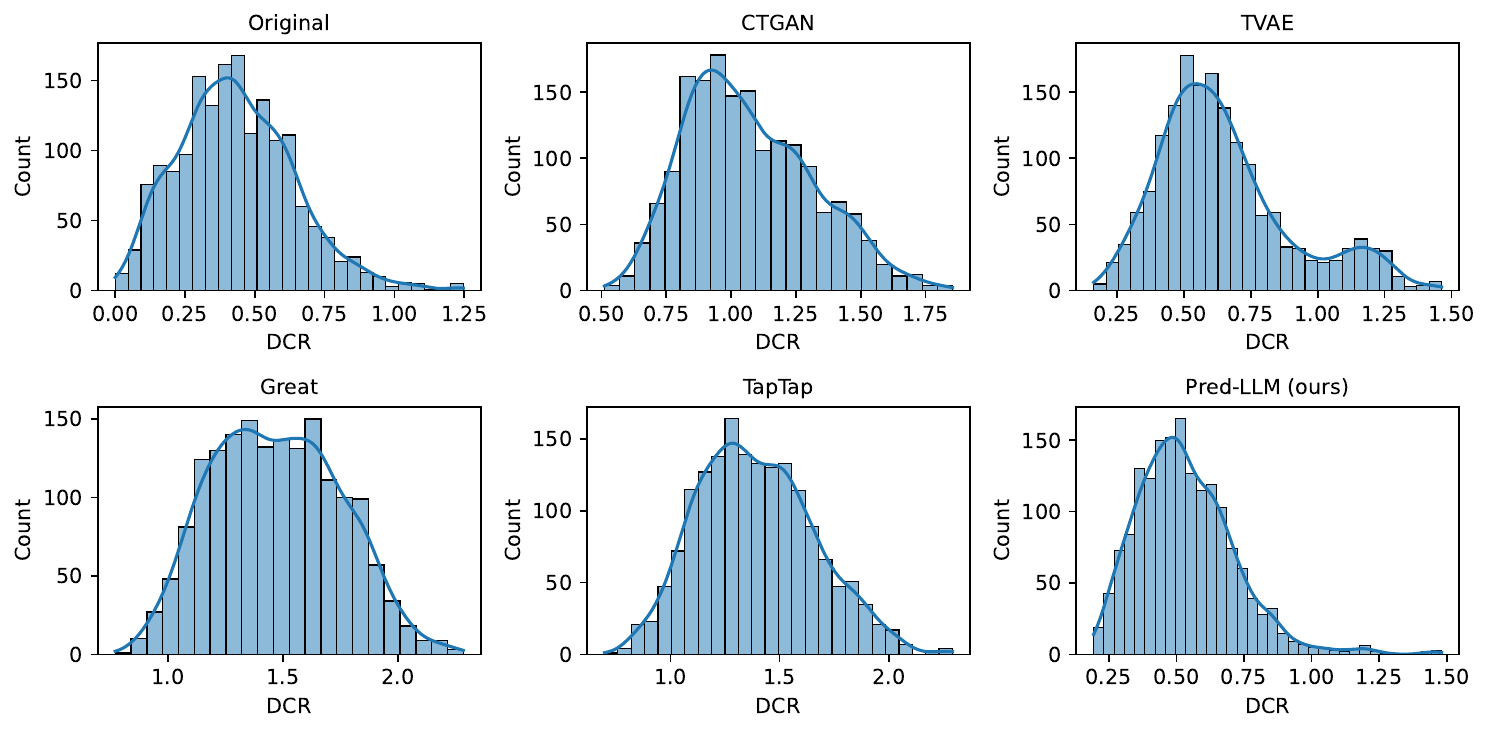}
\par\end{centering}
\caption{\label{fig:DCR-distributions}DCR distributions for the dataset \textit{Cardiotocography}.
The ``Original'' shows the DCR between the real training set ${\cal D}_{real}$
and the real test set ${\cal D}_{test}$. The last plot shows that
our method Pred-LLM does not ``copy'' samples from ${\cal D}_{real}$
but rather synthesizes new samples close to the real samples.}
\end{figure}

Figure \ref{fig:DCR-distributions} visualizes the DCR distributions
of top-5 methods. Only our method Pred-LLM and TVAE can generate synthetic
samples in close proximity to the real samples. Other methods show
significant differences.

\subsubsection{Effect investigation of training size and generation size}

We investigate the effect of the numbers of real and synthetic samples
on the performance of our method. Recall that we use ${\cal D}_{real}$
as the training data to train tabular generation methods and generate
the same number of synthetic samples as the number of real samples
(i.e. $\mid{\cal D}_{fake}\mid=\mid{\cal D}_{real}\mid$).

\begin{figure*}
\begin{centering}
\subfloat[{First, the size of generation data is set to the full training size
i.e. $\mid{\cal D}_{fake}\mid=\mid{\cal D}_{real}\mid$. Then, the
size of training data is reduced by multiplying the full training
size $\mid{\cal D}_{real}\mid$ with a real number in $[0,1]$.}]{\begin{centering}
\includegraphics[scale=0.45]{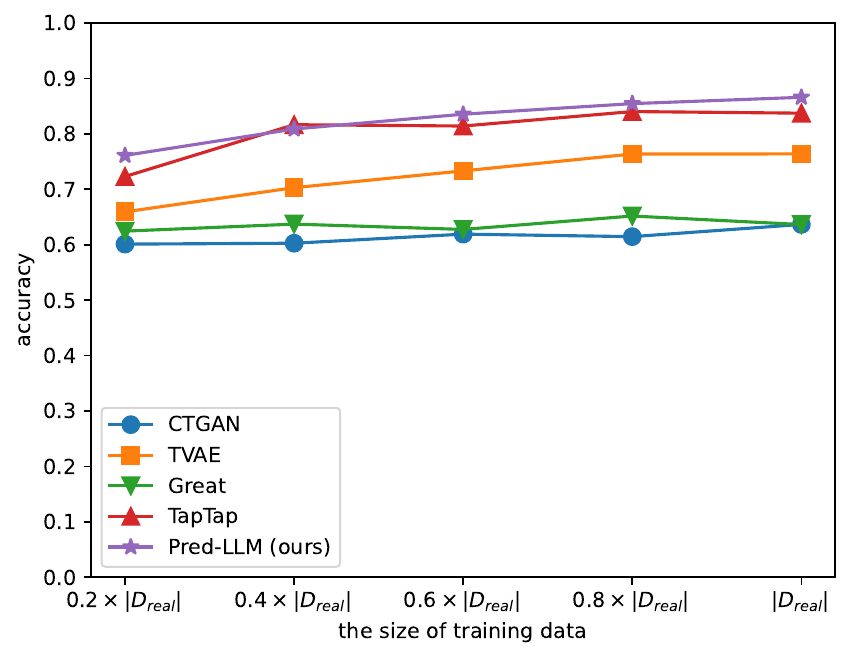}
\par\end{centering}

}\hspace{0.5cm}\subfloat[{First, the training data is full and the size of generation data is
set to the full training size i.e. $\mid{\cal D}_{fake}\mid=\mid{\cal D}_{real}\mid$.
Then, the size of generation data is reduced by multiplying the generation
size $\mid{\cal D}_{fake}\mid$ with a real number in $[0,1]$.}]{\begin{centering}
\includegraphics[scale=0.45]{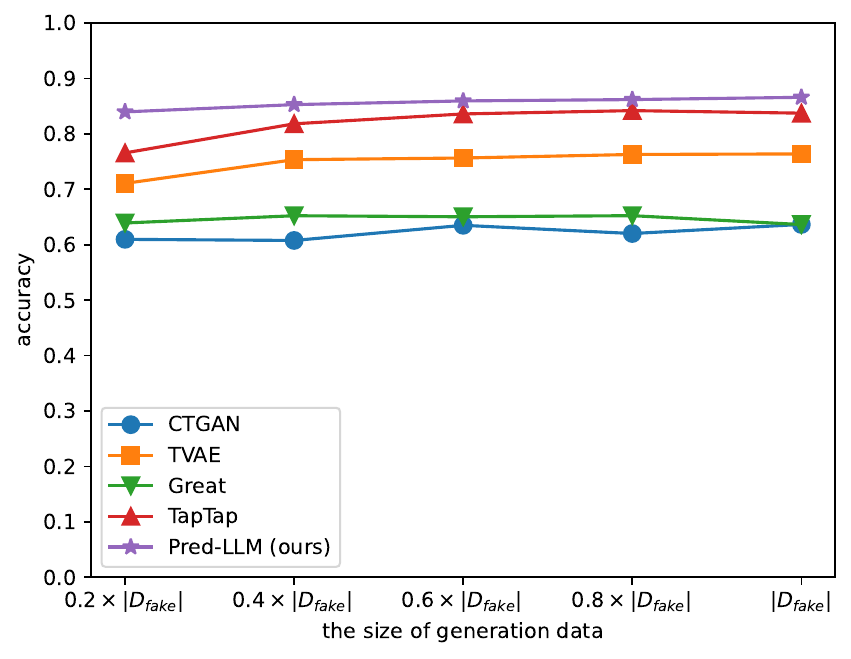}
\par\end{centering}

}
\par\end{centering}
\caption{\label{fig:Train-size-Gen-size}Average accuracy vs. training size
(a) and generation size (b) over 17 classification datasets.}

\end{figure*}

Figure \ref{fig:Train-size-Gen-size}(a) shows our method Pred-LLM
improves when the training size $\mid{\cal D}_{real}\mid$ increases.
It is improved significantly when trained with the full original training
set, improving from 0.7610 to 0.8658. The baseline TapTap also benefits
with more real training samples, improving from 0.7229 to 0.8372.
However, there remains a big gap between its performance and ours.
Similar trends are also observed for other methods.

Figure \ref{fig:Train-size-Gen-size}(b) shows the accuracy when we
train the predictive model XGBoost with different numbers of synthetic
samples $\mid{\cal D}_{fake}\mid$. Here, we still use the full real
dataset $\mid{\cal D}_{real}\mid$ to train tabular generation methods
to generate different cases of $\mid{\cal D}_{fake}\mid$. The classifier
is better when it is trained with more synthetic samples produced
by tabular generation methods except Great. When generating more synthetic
samples, Great causes the classifier's accuracy slightly reduced from
0.6523 down to 0.6363. Our method Pred-LLM makes the classifier better
when it is trained with larger generated data, where the accuracy
improves from 0.8395 to 0.8658.

\begin{figure*}
\begin{centering}
\includegraphics[scale=0.68]{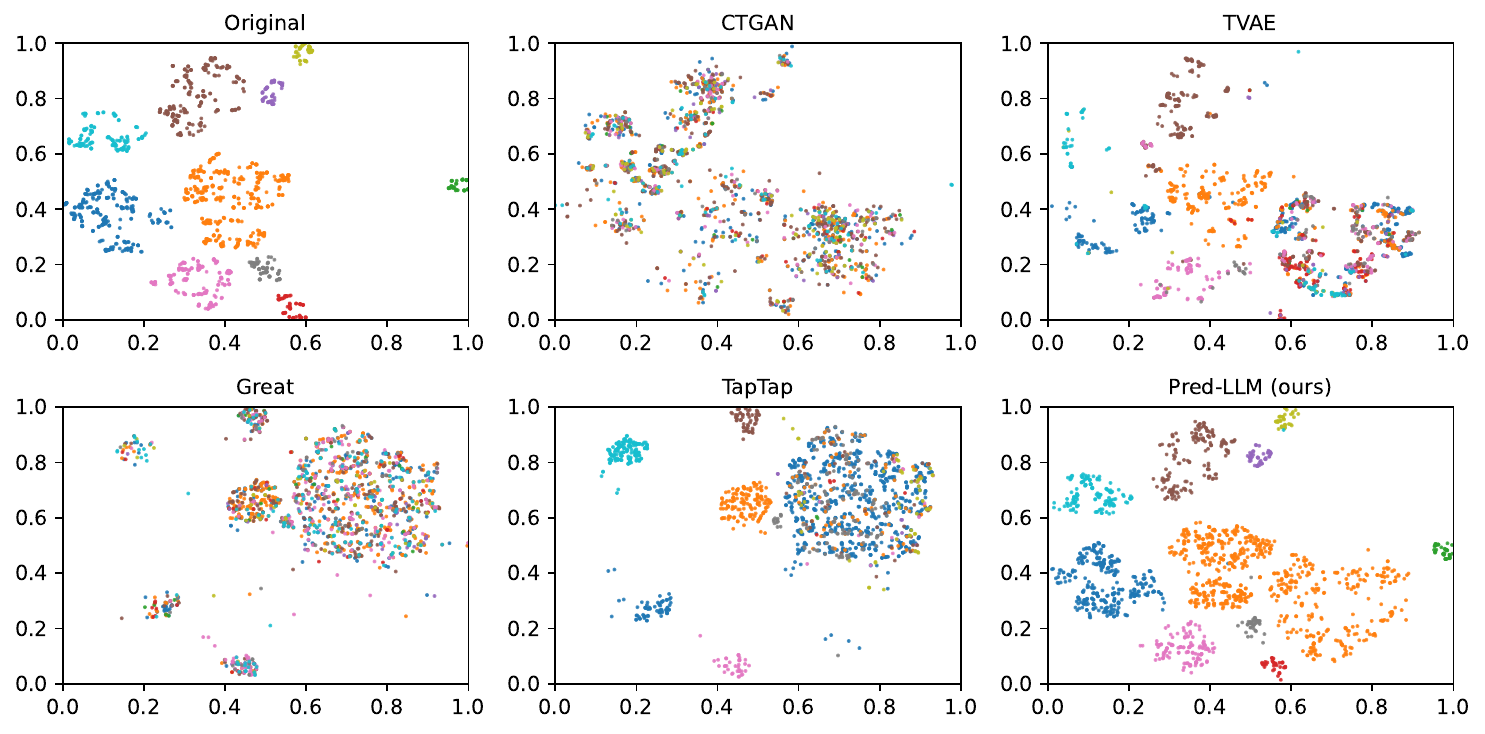}
\par\end{centering}
\caption{\label{fig:Visualization}Visualization of real and synthetic samples
in the dataset \textit{Cardiotocography}. The first plot shows the
\textit{class-wise distribution} of the real samples with 10 classes.
Only our method Pred-LLM can generate synthetic samples close to the
real samples. More importantly, it can capture the true distribution
of the real samples. Other methods generate out-of-distribution synthetic
samples, and they cannot learn the true data distribution.}
\end{figure*}

\subsubsection{Weakness of TapTap}

After generating a synthetic sample $\hat{x}\in{\cal D}_{fake}$,
TapTap uses a classifier trained on the real dataset ${\cal D}_{real}$
(i.e. the Original method) to predict $\hat{y}$, which works well
in most cases. However, when TapTap generates the synthetic samples
whose distribution is \textit{not similar} to that of the real samples,
the classifier may wrongly predict the labels. As shown in Table \ref{tab:TapTap's-weakness},
the Original method has high accuracy on the \textit{real} test set
as $p({\cal D}_{test})=p({\cal D}_{real})$. However, as TapTap cannot
generate \textit{in-distribution} samples i.e. $p({\cal D}_{fake})\neq p({\cal D}_{real})$
(indicated by its low Inverse KL scores), the Original method wrongly
predicts $\hat{y}$ for $\hat{x}\in{\cal D}_{fake}$, leading to a
significant downgrade in TapTap's performance.

\begin{table}[th]
\caption{\label{tab:TapTap's-weakness}TapTap's weakness. As TapTap uses the
classifier trained on ${\cal D}_{real}$ to predict the labels $\hat{y}$
for the synthetic samples $\hat{x}\in{\cal D}_{fake}$, when $p({\cal D}_{fake})$
is significantly different from $p({\cal D}_{real})$ (indicated by
low Inverse KL scores), predicted $\hat{y}$ may be totally wrong,
leading to a poor TapTap.}

\centering{}%
\begin{tabular}{|l|r|r|r|}
\hline 
 & \textbf{Original} & \multicolumn{2}{c|}{\textbf{TapTap}}\tabularnewline
\cline{2-4} 
 & Accuracy & Inverse KL & Accuracy\tabularnewline
\hline 
\hline 
breast & 0.9708 & 0.5652 & 0.8889\tabularnewline
\hline 
steel & 1.0000 & 0.6208 & 0.9135\tabularnewline
\hline 
qsar & 0.8752 & 0.5839 & 0.6619\tabularnewline
\hline 
waveform & 0.8467 & 0.4815 & 0.7473\tabularnewline
\hline 
cardio & 1.0000 & 0.7205 & 0.8998\tabularnewline
\hline 
\end{tabular}
\end{table}

\subsection{Visualization}

For a quantitative evaluation, we use t-SNE \cite{VanderMaaten2008}
to visualize the synthetic samples and compare them with the real
samples.

Figure \ref{fig:Visualization} shows synthetic samples along with
their labels. Here, each color indicates each class. From the real
data (\textit{the 1}\textsuperscript{\textit{st}}\textit{ plot}),
we can see the ``green'' samples lie in the middle of the right
region while the other samples lie in the left region. Our method
Pred-LLM (\textit{the last plot}) is the only method that can capture
this class distribution and generate synthetic samples close to the
real samples. In contrast, other methods generate out-of-distribution
synthetic samples, and they cannot capture the true class distribution
of the real samples.

\section{Conclusion\label{sec:Conclusion}}

In this paper, we address the problem of synthesizing tabular data
by developing a LLM-based method (called Pred-LLM). Different from
existing methods, we propose three important contributions in \textit{fine-tuning},
\textit{sampling}, and \textit{querying} phases. First, we propose
a novel \textit{permutation strategy} during the fine-tuning of a
pre-trained LLM, which helps to capture the correlation between the
features and the target variable. Second, we propose the \textit{feature-conditional
sampling} to generate synthetic samples, where each feature can be
conditioned on iteratively. Finally, instead of leveraging an external
classifier to predict the labels for the generated samples, we \textit{construct
the prompts} based on the generated data to query the labels. Our
method offers significant improvements over 10 SOTA baselines on 20
real-world datasets in terms of downstream predictive tasks, and the
quality and the diversity of synthetic samples. By generating more
synthetic samples, our method can help other applications such as
few-shot knowledge distillation \cite{nguyen2022black} and algorithmic
assurance \cite{Gopakumar2018} with few samples.

\textbf{Acknowledgment:} This research was partially supported by
the Australian Government through the Australian Research Council's
Discovery Projects funding scheme (project DP210102798). The views
expressed herein are those of the authors and are not necessarily
those of the Australian Government or Australian Research Council.

\balance

\bibliographystyle{plain}
\bibliography{reference}

\end{document}